# 모방학습 기반 로봇 작업 속도 개선을 위한 선행 시간적 앙상블 기법

## Proleptic Temporal Ensemble for Improving the Speed of Robot Tasks Generated by Imitation Learning

박 현 준[†] · 임 대 규[1] · 김 승 연[1] · 박 수 민[1]

Hyeonjun Park[†], Daegyu Lim[1], Seungyeon Kim[1], Sumin Park[1]

https://robrosinc.github.io/ACT-PTE/

**Abstract** Imitation learning, which enables robots to learn behaviors from demonstrations by human, has emerged as a promising solution for generating robot motions in such environments. The imitation learning-based robot motion generation method, however, has the drawback of depending on the demonstrator's task execution speed. This paper presents a novel temporal ensemble approach applied to imitation learning algorithms, allowing for execution of future actions. The proposed method leverages existing demonstration data and pre-trained policies, offering the advantages of requiring no additional computation and being easy to implement. The algorithm's performance was validated through real-world experiments involving robotic block color sorting, demonstrating up to 3x increase in task execution speed while maintaining a high success rate compared to the action chunking with transformer method. This study highlights the potential for significantly improving the performance of imitation learning-based policies, which were previously limited by the demonstrator's speed. It is expected to contribute substantially to future advancements in autonomous object manipulation technologies aimed at enhancing productivity.

**Keywords:**   Imitation Learning, Action Chunking, Transformer, Autonomous Object Manipulation

## 1. Introduction

Traditionally, for a robot to autonomously perform a specific task, operators had to manually engineer the robot's movements through rule-based hard coding. This approach requires meticulous reflection of the environment in which the robot operates and demands significant time and expertise to develop the rules[1]. Furthermore, while rule-based behavior generation is useful in structured environments like factories that require high productivity, it has limitations in unstructured environments where the robot's versatility and flexibility are restricted[2].

As robots are increasingly recognized as solutions to various social problems such as low birthrate and labor shortage, there is a growing effort to apply robots in unstructured environments. Imitation learning allows robots to exhibit autonomous behaviors like human behavior by teaching them the desired tasks through demonstrations. These imitation learning algorithms overcome the limitations of robots that only perform limited tasks in structured environments and provide versatility in performing various tasks in dynamically changing environments[3].

Imitation learning works on the principle of which a policy network, trained on data obtained through human's demonstrations, derives the robot's actions based on the observed information. In the field of autonomous driving, for example, driving angles and speeds are determined based on vision data obtained from cameras mounted around the car. In robotics, vision data from cameras and joint data from the robot are used to determine the robot's actions. Specifically, generating actions from image pixels has gained recognition for its potential, as it successfully complete tasks without complex dynamics and environment modeling, leading to an increase in related research[4].

Although the idea of teaching robots through demonstrations has been decades, only recently has the increase in computational power and the advancement of cutting-edge deep learning algorithms such as generative AI and Transformers[5] significantly expanded the range of applicable tasks from laboratory scenarios to real world scenarios. In particular, the Action Chunking Transformers[6] (ACT) has solved issues such as cumulative errors

※ This research was supported by Deep-Tech Tips funded by Ministry of SMEs and Startups
1. Principal Researcher, Robros Inc., Seoul, Korea ({dglim, sy07.kim, mindy} @robros.co.kr)
†. Chief Technology Officer, Corresponding author, Robros Inc., Seoul, Korea (hpark@robros.co.kr)

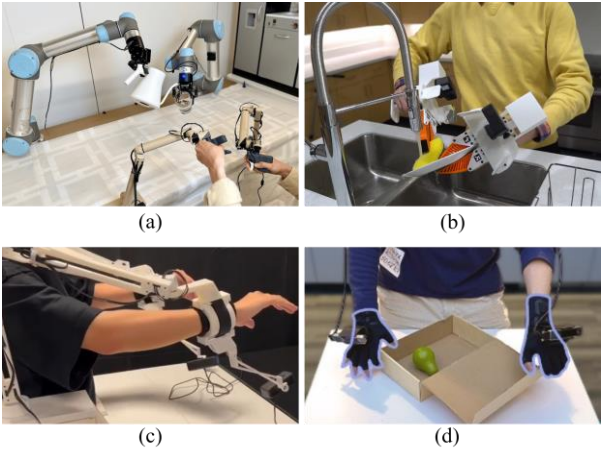

[Fig. 1] Various Data Collection Methods: (a) Remote control of the master device, (b) Portable gripper, (c) Wearable device and hand recognition sensor, (d) Data glove.

in learned policies and problems caused by low-quality data in demonstrations, enabling robots to perform difficult real-world tasks consistently, such as opening a translucent seasoning cup or placing a battery into a slot, with a success rate of 80-90%.

The ACT algorithm operates as follows. First, when the trained policy derives actions from observations, it generates not just a single-step action but an action sequence for the upcoming time horizon. Then, during the stage of deciding the current action, the algorithm combines the predicted future actions at the current time step from the past inferences. This temporal action ensemble significantly enhances the robot's autonomous task performance, but there is still a dependency where the robot's operation speed is constrained by the demonstration data collected through teleoperation.

In this paper, we propose a proleptic temporal ensemble(PTE) algorithm that enables robots to autonomously perform object manipulation tasks faster than the demonstrator's task speed. The proposed method does not incur additional costs as it does not require additional training or computational burden, and it is simple and easy to implement while performing the task three times faster than the existing methods.

This paper is structured as follows. Section 2 describes the imitation learning algorithm based on action chunking, which predicts future action sequences. Section 3 introduces the proposed PTE method. In Section 4, we analyze the performance of the proposed algorithm based on experimental results, and finally, in Section 5, we present the conclusion.

## 2. Behavior Generation of Imitation Learning

### 2.1 Imitation Learning for Dual-Arm Manipulation

Imitation learning (behavior cloning) trains a model to mimic expert behavior by learning from demonstration data that captures actions based on observations. When a demonstrator uses a teleoperation device or other control mechanism to repetitively perform specific tasks with a robot, observations such as images and joint angles, and action data such as desired joint angles are collected during this process. During the training process, the model receives a dataset of observation-action pairs and learns a function that maps the current observation to the demonstrator's future actions.

There are several methods for collecting demonstrator action data. One common approach involves the demonstrator teleoperating the robot using a master arms[6], while another method uses dedicated data collection devices[7]. Additionally, research has explored the use of wearable master arms equipped with cameras that recognize operator's hand movements[8], or directly collecting data by having the demonstrator wear data gloves and body camera designed for data acquisition[9]. [Fig. 1] illustrates these data collection systems. Cameras used for collecting environmental data are typically mounted on the robot's head or attached to the wrist to observe object manipulation. The speed at which the demonstrator performs tasks during data collection directly influences the actions generated by the learned policy.

Various studies have demonstrated successful applications of imitation learning in object manipulation tasks, such as cooking[10] or tying shoelaces[11]. Notable research includes works using Diffusion Policy and Action Chunking with Transformers (ACT), which utilizes Conditional Variational Autoencoders (CVAE)[12]. Both the Diffusion Policy and ACT focus on predicting future action sequences rather than simple immediate actions, operating in a manner similar to Model Predictive Control (MPC), which is called action chunking. Additionally, introducing generative AI techniques such as diffusion models or CVAE model enables the imitation learning model multimodal behaviors of human operators.

In efforts to improve task performance and stability, C. Chi, et al[7] decreases the update speed of the inference by 0.5 times to slow down the execution of the generated behaviors. In these studies, data collected by humans, as shown in [Fig. 1(b)], exhibited fast speeds for the robot causing difficulty in tracking target points. However, increasing the update speed of the network inference for the faster motion of the robot comes with limitations, as it decreases the computational margin for the network inference which takes majority of the computational time due to its large model size.

### 2.2 Action Chunking and Ensemble

In this paper, we adopt the ACT algorithm for imitation learning. ACT fundamentally follows a CVAE structure, with the encoder and decoder using transformers. The CVAE encoder receives the robot's action sequence and current pose as inputs and encodes the robot's action modality into a style vector. The CVAE decoder takes the encoded style vector along with image data acquired through RGB stereo cameras and the robot's end-effector pose sequence as inputs to output the robot's future action sequence. The action space is also the robot's end-effector pose. Defining the observation sequence containing the robot's images and poses from the past time step $t - k_s$ to the current time step $t$ as $s_{t-k_s:t}$, and the action sequence from the current time step $t$ to

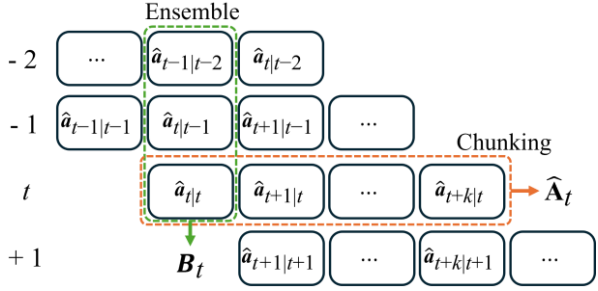

[Fig. 2] Action Chunking and ensemble.

$t + k_a$ as a $a_{t:t+k}$, the policy can be expressed as $\pi_\theta(a_{t:t+k_a} \mid s_{t-k_s:t})$.

[Fig. 2] illustrates the future action sequences predicted by the ACT at each step. The vector $\hat{a}_{u|v}$ represents the estimated action at time $u$ estimated at time $v$, with its magnitude defined by the degrees of freedom of the target robot. When predicting the future action sequences at each step, the action sequence predicted at instance $t$ is as follows

$$\hat{A}_t = [\hat{a}_{t|t} \ \hat{a}_{t+1|t} \ \hat{a}_{t+2|t} \cdots \hat{a}_{t+k_a|t}]. \quad (1)$$

This method of predicting not only the robot's next step action but also the action sequences up to $t + k_a$ is defined as action chunking ([Fig. 1]). Action chunking was designed to address the compounding errors that arise in the robot's intricate object manipulation tasks. Compared to traditional approaches that only infer the next step's action, the use of the action chunking technique enhances the consistency of the robot's behavior. In the ACT algorithm, when applying the action $\hat{A}_t$ to the robot, a weight function is used to ensemble the actions predicted at the current point from the past. First, we define the matrix $B_t$, which consists of the actions inferred from the past to the present, as follows ([Fig. 1]).

$$B_t = [\hat{a}_{t|t} \ \hat{a}_{t|t-1} \ \hat{a}_{t|t-2} \cdots \hat{a}_{t|t-k_a}]. \quad (2)$$

Furthermore, the final input $a_t$ to the robot, incorporating the weight function based on the exponential function, is defined as follows

$$a_t = \sum_i w_i B_t[i] / \sum_i w_i \ , \ w_i = \exp(-m * i). \quad (3)$$

Where the parameter $m$ represents the slope of the weight function. A higher value of $m$ places greater emphasis on the most recently estimated actions. A large value of $m$ enables the robot to respond quickly to unexpected disturbances or environmental changes, resulting in high responsiveness. Conversely, a lower $m$ generates more consistent behavior in the presence of environmental noise. Therefore, it is important to adjust the value of $m$ to find an appropriate balance between the robot's responsiveness and consistency. In all experiments conducted in this study, $m$ was set to 0.05.

### 2.3 The speed limitations of imitation learning policies

The movements of the robot derived from imitation learning policies are entirely dependent on the collected data used during the training process. In the data collection process, the operator manipulates the robot using a teleoperation device, and various task speeds are recorded depending on the operator's skill level. Even if only fast task data is collected from skilled operators to improve the robot's task speed, the robot's task speed based on the imitation learning policy is still constrained by data dependency. The next chapter introduces a proleptic temporal ensemble technique that improves the robot's task speed during the inference process without requiring additional training.

## 3. Proleptic Temporal Ensemble

### 3.1 Proleptic Temporal Ensemble

To accelerate the speed of robot movements generated by imitation learning algorithms, we propose the PTE technique. [Fig. 3] serves as an example to illustrate the proposed method, assuming that five actions are predicted at each step, and displays the actions predicted over the last six steps in blocks. The blocks marked as 'α' in [Fig. 3] represent the actions used at the current time, predicted from the past, and are the components defined in Equation (2) as $B_t$.

The proposed PTE uses $B_{t+f}$ ($f < k_a$) instead of $B_t$ in the calculation of input $a_t$ as described in Equation (3). $B_{t+f}$ is defined as follows

$$B_{t+f} = [\hat{a}_{t+f|t} \ \hat{a}_{t+f|t-1} \cdots \hat{a}_{t+f|t-k_a+f}]. \quad (4)$$

It consists of $f$ fewer $\hat{a}_{u|v}$ compared to $B_t$. For example, if $f$ is 2, $B_{t+2}$ is defined by the blocks marked as 'γ' in [Fig. 3].

As a result, the PTE expects to enhance the robot's task speed by using a weighted average of future control inputs through the ensemble method, instead of relying on the inferred current control input as the target value at the present time.

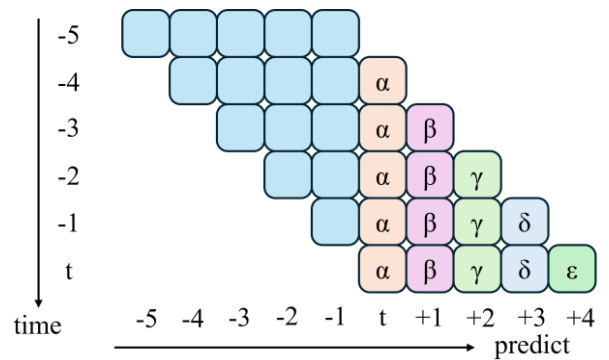

[Fig. 3] Structure of action chunking.

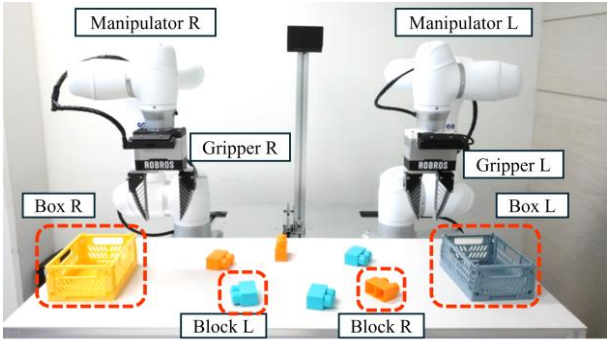

[Fig. 4] Experimental environment.

### 3.2 Proleptic Variables

The proposed PTE technique sets future action points as targets for the robot, allowing it to move more quickly to follow these targets. Research[7] has shown that reducing the action update cycle to increase task speed is not suitable for object contact tasks that require precision at high speeds. Therefore, the proleptic variable *f* needs to be set to an appropriate value depending on the task. Precise tasks require careful execution, while non-contact free motions are better suited for faster speeds. The next chapter validates the changes in task success rates and task speeds based on the proleptic variable *f* through repeated experiments.

## 4. Experimental verification

### 4.1 Block Color Classification Task

To analyze the performance of the algorithm presented in this paper, we selected a block color classification task as the robot's operation. [Fig. 4] illustrates the classification task environment. Two six-degree-of-freedom robotic arms are each equipped with a one-degree-of-freedom gripper to perform the task of picking up yellow and blue 2-stud blocks and placing them into boxes of the corresponding colors. The blocks are identical in size and shape, differing only in color, and their orientations and positions are

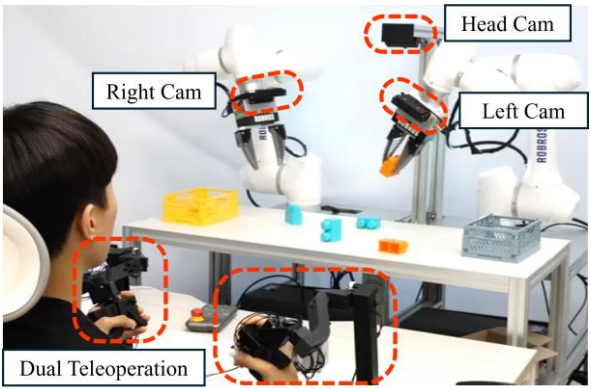

[Fig. 5] Teleoperation environment.

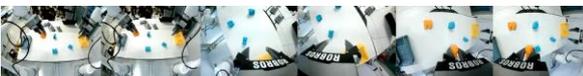

[Fig. 6] Stereo camera images

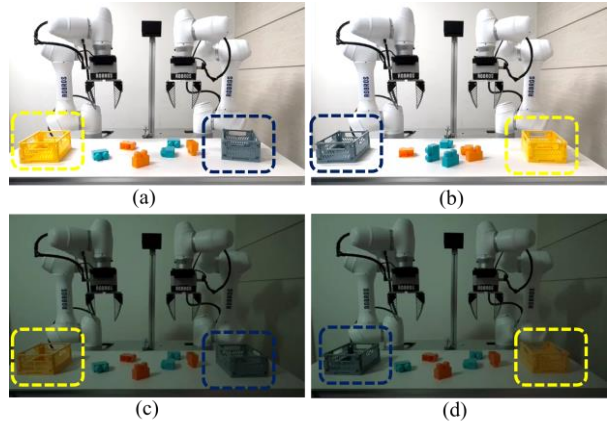

[Fig. 7] Data collecting environment: (a) Basic, (b) boxes swapped, (c) with lights off, (d) with lights off and boxes swapped.

randomly arranged. The boxes are placed at either end of the table without any fixation.

### 4.2 Data Collection for Learning

[Fig. 5] shows the data collection environment. To remotely control the robots, we developed a master control device with a total of seven degrees of freedom, including six degrees for the robot joints and one for the gripper. The master device was designed to have a scale size of 2:1 relative to the robotic arm, and the data structure transmitted from the master device to the robot is as follows

$$[\boldsymbol{p}_r \ \boldsymbol{o}_r \ \theta_r \ \boldsymbol{p}_l \ \boldsymbol{o}_l \ \theta_l]^\mathrm{T}. \tag{5}$$

Where, $\boldsymbol{p}_r \in \mathbb{R}^{3\times 1}$, $\boldsymbol{o}_r \in \mathbb{R}^{3\times 1}$, $\theta_r \in [0,1]$ represent the position of the endpoint of the right arm master device, the Euler angles corresponding to its orientation, and the gripping degree of the gripper, respectively. The subscripts r and l denote the right and left arms, respectively.

Surrounding environmental data was collected using three stereo cameras. As shown in [Fig. 5], two cameras were attached to the wrist, while the remaining one was fixed at a third position. The images captured by the stereo cameras are displayed in [Fig. 6], resulting in a total of six images with a resolution of 640x480.

The demonstrator operated the robot using a teleoperation device, repeatedly performing the task of picking up a total of six blocks and placing them into boxes of the corresponding colors. The blocks were randomly arranged in the center of the table, as illustrated in [Fig. 4], [Fig. 5], and [Fig. 7]. If the box position to which a block needed to be moved during the classification process was outside the robotic arm's workspace, the task was carried out by transferring the block between the robots. The snapshots in [Fig. 11] (e)-(h) illustrate bimanual collaboration in the block sorting task. The right arm overcomes workspace limitations by passing the blue block to the left arm, which then places it in the blue box on the left side.

Since images constitute a significant portion of the data, various lighting conditions were used to collect data, as seen in [Fig. 7]. The

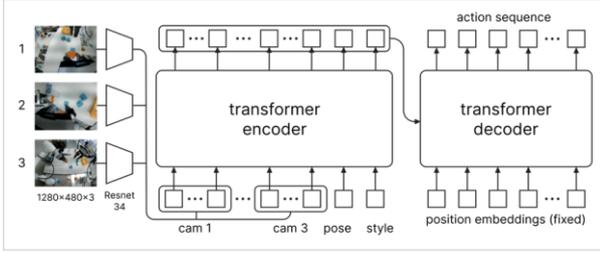

[Fig. 8] Architecture of action chunking with transformer.

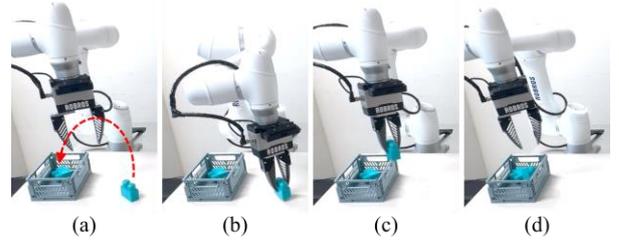

[Fig. 9] Performance Evaluation Experimental Procedure: (a) Initial, (b) Grasp, (c) Carry, (d) Release.

colors of the boxes were also alternated to obtain diverse data. For each environment, data was collected through a total of 1,000 demonstrations, with 250 repetitions per environment, resulting in a total dataset of 1.1 million samples collected at a rate of 20 samples per second.

### 4.3 Policy Learning

[Fig. 8] illustrates the policy architecture utilizing a transformer-based encoder and decoder. Stereo images with a resolution of 640 x 480 are combined in pairs to create three images of size 1280 x 480, which are then input into the encoder. The term 'pose' is defined as the robot's information as follows:

$$[\boldsymbol{p}_r\ \boldsymbol{i}_r\ \boldsymbol{j}_r\ \boldsymbol{k}_r\ \theta_r\ \boldsymbol{p}_l\ \boldsymbol{i}_l\ \boldsymbol{j}_l\ \boldsymbol{k}_l\ \theta_l]^T\ . \quad (6)$$

The unit vectors $\boldsymbol{i}, \boldsymbol{j}, \boldsymbol{k}$ convey the orientation information of each robotic arm's endpoint, representing the columns of the rotation matrices for the right and left arms as follows:

$$\boldsymbol{R}_r = [\boldsymbol{i}_r\ \boldsymbol{j}_r\ \boldsymbol{k}_r],\ \boldsymbol{R}_l = [\boldsymbol{i}_l\ \boldsymbol{j}_l\ \boldsymbol{k}_l]\ . \quad (7)$$

The training was conducted using eight GTX 4090 servers with NVIDIA's Apex library for distributed learning, utilizing a total of 1.1 million data samples over approximately one week and 100,000 training steps. The batch size was set to 256, and the AdamW optimizer was employed as the parameter update algorithm. The policy was structured as $\pi_\theta(\boldsymbol{a}_{t:t+24}\ |\ \boldsymbol{s}_{t-1:t})$ where it receives the latest two environmental inputs to predict the future sequence of 24 actions. The update frequency for the robot's target point was set to 20 Hz.

### 4.4 Evaluation of PTE Performance

In this paper, we conducted four types of experiments. The first experiment is a four-block sorting task to validate the effectiveness of the ACT-based imitation learning policy. The second experiment is a repetitive single-block moving task to evaluate the speed performance of PTE. The third experiment is an unseen object sorting task to assess the scalability of the policy. The fourth is a four-block sorting task to validate the feasibility of PTE-ACT.

First, to evaluate the effectiveness of the ACT-based imitation learning policy, we conducted a four-block sorting experiment. The experiment was implemented with an *f* value of 0 and was executed a total of 60 times. The experimental environment was set up by randomly scattering the positions and orientations of two blue blocks and two yellow blocks, as well as the location of the box, under varying lighting conditions.

The success rate for the classification task was recorded at 65%. The failure cases included 12% due to color classification errors, 18% attributed to covariate shift problems, and 5% where blocks were ejected during grasping. The covariate shift problem refers to errors occurring in state spaces that were not addressed by the collected data. For example, there were cases where two robotic arms attempted to grasp a block positioned in the center simultaneously, leading to a halt to avoid a collision, or situations where they failed to recover their posture under specific conditions.

The second experiment is a single-block sorting task to evaluate the speed performance of PTE. In case of multi-block sorting tasks, the variance in elapsed time is high due to the number of block handovers between robots, which depends on the initial block arrangement. Therefore, to objectively evaluate the speed performance of PTE, we performed a repetitive single-block sorting task. Experiments were conducted to pick up one block and place it into the nearest box for various values of *f*. [Fig. 9] illustrates the experimental process. For each value of *f*, 20 trials were conducted, and the average time taken and success rates were listed in [Table 1]. Instances where the block was dropped during the picking process or not placed inside the box were considered failures. The experimental results demonstrated that PTE improved the task speed by approximately 2 times while maintaining a success rate of 100%. Additionally, although the success rate decreased to 75%, it demonstrated a threefold improvement in task speed.

### 4.5 Discussion and Future Work

The third experiment is an unseen object sorting task to evaluate the scalability of the policy. In the data collection process, only

Table 1. Experiment result with various *f*.

| *f* | Average elapsed time | Succes rate |
|---|---|---|
| 0 | 7.274 | 100 |
| 5 | 5.480 | 100 |
| 10 | 4.024 | 100 |
| 15 | 3.214 | 95 |
| 20 | 2.418 | 75 |

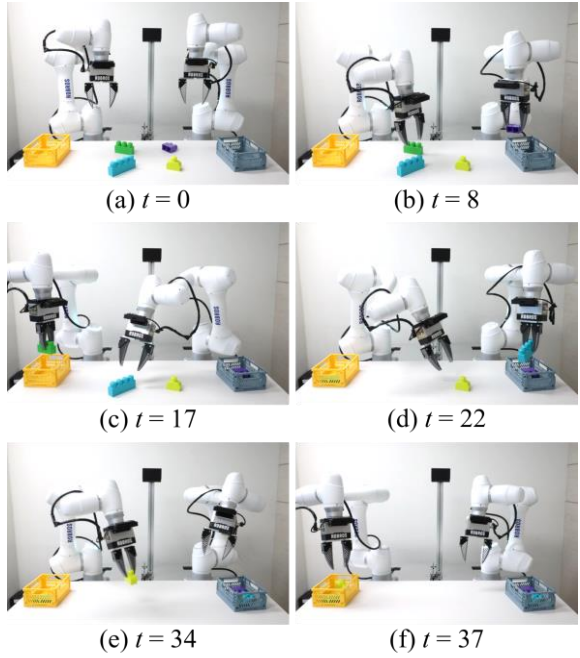

[Fig. 10] Snapshots of the object classification task for unseen objects during the training process with $f = 0$.

yellow and blue 2-stud blocks were utilized. To investigate the scalability of the policy, classification experiments were conducted on previously uncollected data for blue 4-stud block, green 3-stud block, purple 2-stud block, and differently shaped yellow-green blocks [Fig. 10]. The experimental results indicated that the blue and purple blocks were classified into the blue box, while the green and yellow-green blocks were classified into the yellow box.

The final experiment is a four-block sorting task to comprehensively evaluate the performance of PTE as shwon in [Fig. 11]. PTE-ACT took only 13 seconds to sort four blocks, including block handovers. This contrasts with the 37 seconds required by ACT alone to sort four blocks, as shown in [Fig. 10].

For environmental perception, this study employed three stereo cameras. The wrist camera was used to observe the gripping state of the objects, while the head camera was utilized to monitor the color information of the boxes placed on both sides. Let us assume a block classification task conducted using the policy $\pi_\theta(a_{t:t+k_a} \mid s_{t-k_s:t})$ with a low $k_s$. [Fig. 11](f) depicts a scene where a block is handed over to the left, but when viewed in isolation, it could be interpreted as the block being received from the right. If only the wrist camera were used for the classification task, there would be no color information regarding both boxes, resulting in observations of the two robots repeatedly handing blocks to each other. In future work, we plan to conduct research on generating appropriate actions when the acquired environmental information is the same, but the situations differ.

This paper proposes the PTE algorithm for improving the task speed of robot actions generated by imitation learning policies. Upon examining the collected data, the four-block sorting task performed by the teleoperator took an average of 34 seconds. Successful cases of the autonomous block sorting task using only ACT elapsed an average of 41 seconds. The PTE-ACT reduced

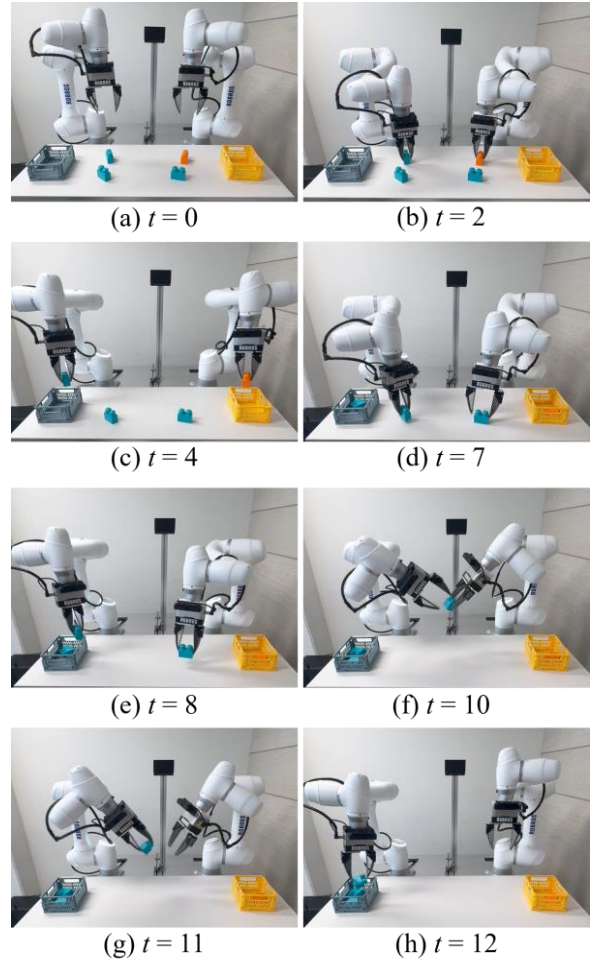

[Fig. 11] Snapshots of the object classification task with $f = 19$.

the time to 12 seconds, demonstrating a performance that is approximately 3 times faster than human teleoperation tasks. However, PTE showed a tendency to decrease stability in dexterous object manipulation tasks, which led to a lower success rate. Future work will focus on research to improve both speed and success rate simultaneously.

## 5. Conclusion

In this paper, we propose a PTE that enhances the performance of imitation learning algorithms, which generate robot actions by learning from demonstrator data. The proposed algorithm operates by incorporating future actions predicted by the policy into the current stage to improve the speed of robot tasks. Our proposed method has the advantage of saving time and costs as it does not require new data collection or policy retraining. To evaluate the performance of the proposed PTE algorithm, we conducted a block color classification task using a CVAE-based ACT algorithm with a dual-arm robot. The experimental results confirmed a 3x improvement in speed. This study demonstrates the enhanced performance of imitation learning-based policies, which were previously dependent on the demonstrator's task speed. It is expected to contribute to productivity increases in future autonomous object manipulation technologies.

## Acknowledgments

We thank to Ohyun Kwon and Dongsu Yang for their contributions to data collection and support.

## References

[1] T. Lozano-Perez, "Robot programming," *Proceedings of the IEEE, Vol. 71, No. 7, 1983.*

[2] G. Biggs and B. MacDonald, "A survey of robot programming systems," *Proceedings of the Australasian conference on robotics and automation*, pp. 1-3, 2003.

[3] A. Hussein, et al., "Imitation learning: A survey of learning methods." *ACM Computing Surveys*, Vol. 50, No. 2, pp. 1-35, 2017.

[4] M. Zare, et al., "A survey of imitation learning: Algorithms, recent developments, and challenges." *IEEE Transactions on Cybernetics, pp.1-14,* 2024.

[5] A. Vaswani, "Attention is all you need." *Advances in Neural Information Processing Systems,* 2017.

[6] TZ. Zhao, et al., "Learning fine-grained bimanual manipulation with low-cost hardware." *arXiv preprint arXiv:2304.13705,* 2023.

[7] C. Chi, et al., "Universal manipulation interface: In-the-wild robot teaching without in-the-wild robots." *arXiv preprint arXiv:2402.10329,* 2024.

[8] S. Yang, et al., "Ace: A cross-platform visual-exoskeletons system for low-cost dexterous teleoperation." *arXiv preprint arXiv:2408.11805,* 2024.

[9] C. Wang, et al., "Dexcap: Scalable and portable mocap data collection system for dexterous manipulation." *arXiv preprint arXiv:2403.07788,* 2024.

[10] Fu, Zipeng, Tony Z. Zhao, and Chelsea Finn. "Mobile aloha: Learning bimanual mobile manipulation with low-cost whole-body teleoperation." arXiv preprint arXiv:2401.02117 (2024).

[11] Zhao, Tony Z., et al. "Aloha unleashed: A simple recipe for robot dexterity." arXiv preprint arXiv:2410.13126 (2024).

[12] Chi, Cheng, et al. "Diffusion policy: Visuomotor policy learning via action diffusion." *The International Journal of Robotics Research* (2023): 02783649241273668.

## 저자 약력

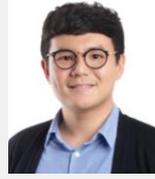

**박 현 준**

2011 광운대학교 제어공학과 (공학사)
2014 서울대학교 융합과학기술대학원 지능형융합시스템학과 (공학석사)
2020 서울대학교 융합과학기술대학원 지능형융합시스템전공 (공학박사)

2020~2021 한국로봇융합연구원 선임연구원
2021~현재 (주)로브로스 이사
관심분야: 휴머노이드, 모방학습, 로봇 핸드, 물체 조작

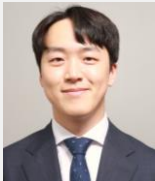

**임 대 규**

2017 서울대학교 기계항공공학부 (공학사)
2024 서울대학교 융합과학기술대학원 지능형융합시스템전공 (공학박사)
2024~현재 (주)로브로스 책임연구원

관심분야: 휴머노이드, 모방학습, 강화학습, 딥러닝

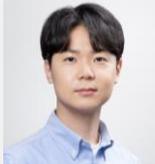

**김 승 연**

2015 서울대학교 기계항공공학부 (공학사)
2023 서울대학교 융합과학기술대학원 지능형융합시스템전공 (공학박사)
2023~2024 삼성전자 종합기술원

2024~현재 (주)로브로스 책임연구원
관심분야: 휴머노이드, 로봇핸드

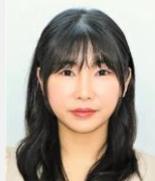

**박 수 민**

2010 서울과학기술대학교 기계설계자동화공학부 (공학사)
2012 서울대학교 융합과학기술대학원 지능형융합시스템학과 (공학석사)
2020 서울대학교 융합과학기술대학원 지능형융합시스템전공 (공학박사)
2020~2021 한국생산기술연구원 박사후연구원

2021~현재 (주)로브로스 책임연구원
관심분야: 대규모언어모델, 휴머노이드 로봇 보행, 생체역학